\documentclass[letterpaper]{article} 
\usepackage{aaai24}  
\usepackage{times}  
\usepackage{helvet}  
\usepackage{courier}  
\usepackage[hyphens]{url}  
\usepackage{graphicx} 
\usepackage{natbib}  
\usepackage{caption} 
\usepackage{amsmath, amssymb}
\usepackage{mathtools}
\usepackage[capitalise]{cleveref}
\DeclareCaptionStyle{ruled}{labelfont=normalfont,labelsep=colon,strut=off} 
\usepackage{algorithm}
\usepackage{algorithmic}
\usepackage{booktabs}       
\usepackage{todonotes}
\usepackage{siunitx}
\usepackage{comment}
\usepackage{titling}
\usepackage{newfloat}
\usepackage{listings}
\usepackage{xspace}
\usepackage{subcaption}
\floatstyle{ruled}
\newfloat{listing}{tb}{lst}{}
\floatname{listing}{Listing}

\newcommand{\expec}[1]{\mathbb{E}[#1]}
\newcommand{\bsdem}[0]{q_{it}^{(b)}}
\newcommand{\DML}{{DML}\xspace}
\newcommand{\DMLForecaster}{{DML Forecaster}\xspace}
\newcommand{\Baseline}{{TF}\xspace}
\usepackage{authblk}

\usepackage{tabularx}

\title{Causal Forecasting for Pricing}

\author[1]{Douglas Schultz}
\author[1]{Johannes Stephan}
\author[1]{Julian Sieber}
\author[1]{Trudie Yeh}
\author[1]{Manuel Kunz}
\author[1]{Patrick Doupe}
\author[1]{Tim Januschowski}

\affil[1]{Zalando SE}
\affil[ ]{\small{{\{douglas.schultz, johannes.stephan, julian.sieber, trudie.yeh, manuel.kunz, patrick.doupe, tim.januschowski\}@zalando.de}}}

\usepackage{bibentry}

\begin{document}
\maketitle

\begin{abstract}
This paper proposes a novel method for demand forecasting in a pricing context. Here, modeling the causal relationship between price as an input variable to demand is crucial because retailers aim to set prices in a (profit) optimal manner in a downstream decision making problem. Our methods bring together the Double Machine Learning methodology for causal inference and state-of-the-art transformer-based forecasting models. In extensive empirical experiments, we show on the one hand that our method estimates the causal effect better in a fully controlled setting via synthetic, yet realistic data. On the other hand, we demonstrate on real-world data that our method outperforms forecasting methods in off-policy settings (i.e., when there's a change in the pricing policy) while only slightly trailing in the on-policy setting.
\end{abstract}

\section{Introduction}
Time series forecasting in practical applications commonly feeds into decision problems in multiple domains~\citep{petropoulos2021forecasting}. 
We consider the special case of an online fashion retailer, where demand forecasts play a key role in setting optimal prices for a large collection of articles~\citet{li2021large}. 
Our task consists of predicting demand subject to different price levels or discounts which the online retailer controls at least partially.\footnote{There is also a competitive component in pricing that we ignore in the context of this work.} 

Our use case requires two types of estimates to make pricing decisions. First, we need to predict demand at different prices for multiple weeks in the future. In online retail contexts the focus is mainly on predicting demand levels~\citep{seeger2016bayesian,wen2017,kunz2023deep}. Second, we need to understand the causal effect of price changes on demand to choose among price levels. The \textit{price elasticity of demand} is the percentage change in demand for a percentage change in price. An elasticity is useful in setting prices as in simple cases an elasticity can be used with marginal costs to set optimal prices~\citep{phillips2021pricing}. Our use case is more complex. We also need the forecasted level of demand at different prices for multiple weeks in the future. So we need to combine forecasts with causal inference to make good pricing decisions.

Our paper bridges the gap between forecasting and causal inference in the context of demand forecasting for pricing. We take an opinionated approach in the sense that predictive accuracy is what we focus on, but the model we present here heavily leans on causal inference machinery in particular the Double Machine Learning (\DML) framework~\citep{chern2017}. 
Our contributions are as follows:
\begin{itemize}
\item We present a novel forecasting modeling framework using the classic \DML split into an outcome model, a treatment model and an effect model. For each model, we use state-of-the-art transformer based models.
\item We design \& provide synthetic, but realistic data for empirical evaluations in a fully-controlled environment on the one hand, and show on the other hand, how real-world data can be used in counterfactual scenarios for effective evaluation via commonly occurring natural experiments or how to mimic them effectively.
\end{itemize}
In empirical evaluations, we show that our model performs roughly on par with state-of-the-art forecasting models in a standard, on-policy setting, but has a clear advantage in off-policy settings where the forecast horizon contains price policies that haven't been observed in the training set. 

Our paper is structured as follows. We formalize the problem setting in~\cref{sec:problem_setting}. We present our model in~\cref{sec:model} and evaluate it in~\cref{sec:experiments} on both synthetic, open source data sets and  a real-world, closed source data set. We discuss related work in ~\cref{sec:related_work} and conclude in~\cref{sec:conclusion}.

\section{Problem Setting and Background}\label{sec:problem_setting}

For any time series $x$, $x_{0:T}$ is short-hand for $[x_0, x_1, \ldots, x_T]$. 
The observational time series data of an article $i$ at time $t$ starting at $0$ is given by $\{q_{i0:t}, d_{i0:t}, z_{i0:t}\}$, where 
$q$ denotes the demand, $d$ corresponds to the discount, which is the percentage of price reduction relative to the article's recommended retailer price; and $z$ a set of article specific covariates. These can include past demand in particular, but also time-independent variables such as catalog information. The object of interest is 
\begin{equation}\label{eq:demand_model_prob_causal}
P(q_{it+1:t+h}| \mathrm{do} (d_{t+1:t+h}), q_{0:t}, d_{0:t}, z_{0:t+h}; \theta)\;,
\end{equation}
that is, the probability distribution of demand in the forecast horizon $t+1:t+h$ conditioned on covariates and discounts in the forecast horizon on which we can intervene, hence $\mathrm{do} (d_{t+1:t+h}$).
A standard approach is to simplify~\eqref{eq:demand_model_prob_causal} to a conditional expectation that is estimated via some time series model, without explicitly modeling the effect of interventions.
\begin{equation}\label{eq:current_estimand}
\mathbb{E}[q_{it+1:t+h}| d_{t+1:t+h}, q_{0:t}, d_{0:t}, z_{0:t+h}] \;.
\end{equation}
To model these interventions we often assume conditional ignorability, positivity and consistency~\citep{hernan2010causal,chern2017,cunningham2021causal}. In this work we do not assume these as we're interested in improving our forecasts, not estimating treatment effects. For instance, given the dynamic patterns in the data we might not adjust fully for all confounders and not meet conditional ignorability. Meeting these assumptions will result in unbiased treatment effect estimates and improve estimates.

A standard approach to estimate the effect of an intervention is via 
\DML, which we introduce briefly. While \DML is typically used to estimate binary or discrete
treatment effects~\citep{chern2017}, we take ideas from \DML for estimating the effect of a continuous treatment variable: weekly average discount, with an outcome of demand. As in~\citet{chern2017}, we introduce \DML using a partial linear model:
\begin{eqnarray}
q & = &  d \theta + g(z) + u, \, \, \mathbb{E}[u|z,d] = 0 \\ 
d & = & m(z) + v, \, \, \mathbb{E}[v|z] = 0  \nonumber
\end{eqnarray}
Here our target $q$ (demand) depends on the control input $d$ (discounts), effects of the
environment $z$ and independent noise $u$. $\theta$ is the linear effect of $d$ on our target $q$, and thus the causal parameter of interest. The effect of $z$ on $q$ is passed through the function $g$ that
can adopt any shape. Furthermore, the treatment $d$ is
affected by our environment $z$ via $m$ as well as some independent random
component $v$.

DML undergoes two stages: the
nuisance stage and the effect stage. The nuisance stage includes two
\emph{nuisance models} which predict treatment (discount) and outcome (demand), whereas the latter is computed without using future discount as input. The ground truth treatment and outcome
are then residualized using the predictions of these nuisance models and passed on to the effect model in order to compute a
treatment effect. The final output is then the output of the effect model taken
together with the output of the outcome model and the desired treatment.
Typically, all three of these models are trained separately with separate
losses. Note that, the training of the effect model uses the output of
the nuisance models and therefore requires a special treatment. 

The benefit of orthogonalization is that we account for regularization bias~\citep{chern2017}, which affects S-Learners\footnote{S for Single learner. We can calculate treatment effects by augmenting the treatment feature and subtracting. E.g. $\mathbb{E}[q| d=0.5] - \mathbb{E}[q| d=0.4]$. This is otherwise known as G-computation in epidemiology and other fields.} like~\eqref{eq:current_estimand}. In the standard practice, discounts are treated like any other independent variable and thus regularized/shrunk in order to improve predictions. This regularization biases estimates of the causal effect between discounts and demand~\cite{chern2017}. 

\section{\DMLForecaster}\label{sec:model}
\begin{figure}
    \centering
    \includegraphics[width=0.4\textwidth]{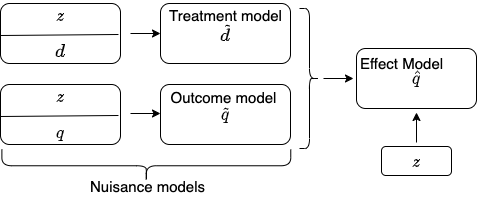}
    \caption{The architecture of the \DMLForecaster.}
    \label{fig:dml_forecaster}
\end{figure}

Our approach for a causal forecaster follows the \DML approach and it hence consists of three submodels that we introduce in the following. 
\cref{fig:dml_forecaster} depicts the high-level architecture. 

\paragraph{The Nuisance Models}

Each of the two nuisance\footnote{We use this term to remain consistent with the causal inference literature; however, the outcome nuisance model is of primary interest for our use case.} models provide estimates $\tilde{q}$ and $\tilde{d}$ of $q$ and $d$ given $z$ respectively. We call the model that provides $\tilde{q}$ the \emph{outcome} model and the model that provides $\tilde{d}$ the \emph{treatment} model. Here, we choose standard transformer-based forecasting models~\citep{vaswani2017} for their robustness and proven performance in an online retail setting~\citep{eisenach2020mqtransformer,rasul2021multivariate,zhou2021informer}.

Each of the outcome and treatment prediction models have the same architecture and only differ in target and final activation. We use \texttt{softplus} as the final activation function in the outcome model to enforce positivity. For our treatment model, we do not pass the (linear) combination learned by the last layer through a (non-linear) activation function. As we will see from the functional form of the effect model head, this
is helpful given the multiplicative nature of the effect model.
Around each attention step, there
is a residual connection, and after each attention step there is a
position-wise feed forward network with layer normalization and dropout.
We use an $L1$ loss for training our nuisance models on real-world data and an $L2$ loss when fitting our synthetic data set. Other choices of losses are possible and our approach readily extends to these, in particular for probabilistic scenarios~\citep{gneiting2007probabilistic}.

\paragraph{The Effect Model}\label{sec:effect_model}
The effect model combines the treatment and outcome models to provide the final estimate of demand $q$ in our \DMLForecaster which we denote as $\hat{q}$. We now show how our model estimates the price elasticity of demand
\begin{equation}\label{elastdef}
\epsilon \coloneqq \dfrac{\Delta q}{q}\cdot \dfrac{p}{\Delta p},
\end{equation}
where $q$ is demand of an article and $p$ is the price. If we assume mild integrability conditions, then basic integration gives us
\begin{equation}\label{eq:elasticity}
q_1=q_0\Big(\frac{p_1}{p_0}\Big)^\epsilon,
\end{equation}
where $q_i$ is the demand at price $p_i$ (see~\cref{app:derivation_demand_eq} for details).

Our idea is to parameterize $\epsilon$ by a neural network. Given retail price $x$, we can write the discounted price as $x\cdot(1-d_t)$ where $d_t$ is the discount at time $t$. Furthermore, we assume that the forecast of the outcome model $\widetilde{q_t}$ is an estimate of the sales at the price level predicted by the discount model $x\cdot(1-\widetilde{d_t})$. Substituting these into \cref{eq:elasticity}, we can compute our final demand estimate $\widehat{q}$ at time $t$ as
\begin{equation}\label{eq:effecthead}
\widehat{q_t}=\widetilde{q_t}\Big(\frac{1-d_t}{1-\widetilde{d_t}}\Big)^{\psi(z)} \qquad 1\leq t \leq s,
\end{equation}
where $\psi$ is a transformer model whose output is the elasticity $\epsilon$ in~\eqref{eq:elasticity} and $s$ is the length of the forecast horizon. Note, that while $\epsilon$ is assumed to be constant here, it still is parameterized over $z$ so it can vary by features used in estimation.
Our model to parameterize $\epsilon$ is similar to the nuisance models and only lacking the decoder self-attention as we expect elasticity to be relatively constant within the forecast horizon. The outcome model accounts for the auto-regressive part of each time series. We use a negative \texttt{softplus} as final activation as we expect elasticity to be negative~\cite[Chapter 15]{varian2014intermediate} and an $L_1$ loss for training.

For training the nuisances and effect models, we deploy a two-stage training process, where we fit the nuisance models in the first stage and the effect model in the second stage. The first stage nuisance models generate estimates for the second stage effect models. 

To avoid overfitting, we deploy two-fold cross-fitting during training in a similar manner to \cite[Section 3]{chern2017}. We have an even and odd copy of each nuisance model, each of which are trained on one half of the data set. We use nuisance models trained on odd data to infer outcomes for even data, and vice-versa. This data is used to train a single effect model.

The splitting of the data set into even and odd parts is done according to the index of the item $i$. In the particular instance of demand forecasting, we can derive an index from article information such that articles of the same size are guaranteed to be either even or odd indexed while still having a (close to) random split between different articles. 

\paragraph{Inference with the \DMLForecaster}

Once the model is trained, we need to infer future outcomes for different discount levels. We combine two methods here, one influenced by the above cross fitting procedure and one influenced by standard forecasting methods. We ensemble these two methods with a geometric mean, where $cf$ indicates cross fitting and $f$ indicates forecasting
\begin{equation}\label{eq:ensemble}
\widehat{q_t} = \sqrt{
\widetilde{q_{t}^{cf}}\Big(\frac{1-d_t}{1-\widetilde{d_{t}^{cf}}}\Big)^{\psi(z)}
\cdot \widetilde{q_{t}^f}\Big(\frac{1-d_t}{1-\widetilde{d_{t}^f}}\Big)^{\psi(z)}
}
\end{equation}
In the cross fitting procedure we pass the odd (even) batches to the even (odd) nuisance models, and then receive an inference from the effect model. We do this to account for potentially overfit models. 

The standard forecasting practice is to use the model trained on old data to infer future outcomes. To implement this we pass even (odd) batches to the even (odd) nuisance models, and then pass the output to the effect model. This has the advantage of forecasting ahead on items that the nuisance models have seen during training. 

\paragraph{Discussion: Departures from the \DML Literature}

We’re interested in forecasting demand levels for different discount rates. The DML literature is interested in estimating changes in demand levels for changes in discount rates. Although we use cross fitting to train our model, at inference we depart from this, for improved forecast performance. Second, we use a single effect model, instead of separate, averaged treatment effect estimations on each half of the dataset. Third, we use an outcome model that reflects our understanding of the problem space, and not one justified for treatment effect estimation.\footnote{We could have assumed an outcome function which depends neither non-linearly nor log-linearly on treatment, and used a learned weighted sum of the raw output of the effect transformer as the final output, with the small modification of also providing the nuisance outputs and true discounts to the effect transformer. Such a model showed similar metrics in preliminary experiments.}

\section{Experiments}\label{sec:experiments}
In this section, we present experimental results of the \DMLForecaster in a fully controlled setting with synthetic data and on real-world data. We start by discussing practical details around the \DMLForecaster.

\subsection{Baseline Models and Accuracy Metrics}

We compare the \DMLForecaster to the following models:
\begin{itemize}
    \item \emph{Na\"ively-causal Transformer (\Baseline):} A time-series transformer architecture with a special output head that models price elasticity more generally than~\eqref{eq:elasticity} via a piece-wise linear, monotone function~\citep{Kunz2023}. 
    
    \item \emph{SARIMAX:} A vanilla seasonal ARIMA model with exogenous covariates. In cases where the training length was less than 30, or the model fitting process failed, we use the previous week's value as a fallback. For our experiments, we use Darts 0.21.0~\citep{darts}, co-variates such as stock and discount variables from previous time steps were included, and preprocessing involved log transformation and forward filled for missing values in demand, stock (in $z$), and discounts.
    
    \item \emph{TWFE elasticities:} A standard econometric baseline via a causally informed, elasticity-based forecast using a two-way fixed-effect Poisson regression model~\cite{fixest}. Appendix~\ref{app:TWFE} contains more details.
    
    \item \emph{sDML:} As part of our ablation study, this model implements the \DMLForecaster (see \cref{sec:effect_model}) without the nuisance model for predicting the treatment. Instead the treatment is provided directly to the effect model without residualization.

    \item \emph{No Cross Fitting:} Cross fitting is applied to the DML Forecaster as described in Section~\ref{sec:effect_model}. For our ablation study, we create variants of sDML and DML models without cross fitting (sDML-no cf and DML-no cf).
\end{itemize}
We have chosen these models to represent the variety of approaches typically deployed for such problems~\citep{JANUSCHOWSKI2020167}: (i) local forecasting models (SARIMAX), (ii) econometric approaches (TWFE) and (iii) global, transformer-based forecasting methods. 

For the accuracy metrics, we use standard metrics mean absolute error (MAE) and mean squared error (MSE)~\citep{hyndman2017forecasting}, and the so-called \emph{demand error}, a metric that captures the down-stream pricing dependency (see~\citet{kunz2023deep}):
\begin{equation}\label{def:demand_error}
    \mathcal{D}_{T,h} = \sqrt{\frac{\sum_{i}\sum_{T=t+1}^{t+h}b_i(\hat{q}_{i,T} - q_{i,T})^2}{\sum_{i}\sum_{T=t+1}^{t+h}b_iq_{i,T}^2}}.
\end{equation}
Here, $t$ is the last timepoint in the training set, $h$ is the forecast horizon, 
$\tilde{q}_{i,T}$ is the prediction for article $i$ at timepoint $T$, $q_{i,T}$ is the corresponding true demand and $b_i$ is
the recommended retail price of article $i$.

\subsection{Hyperparameter Tuning}
The following provides an overview on how we select the hyper-parameters. More details are in \cref{app:hyperparameters}. 
\paragraph{Synthetic Dataset}
We use Bayesian optimization~\citep{akiba2019optuna} to tune key hyperparameters of the \DMLForecaster and \Baseline. To mimic a realistic tuning, we use the data of the first 50 weeks of our simulated data whereas we keep weeks 46-50 as a hold out set to select the best hyperparameters, and thus use the first 45 weeks for training. In the case of \DML-no cf, we reuse the same hyperparameters found for the \DMLForecaster. For sDML and sDML-no cf, we only need to re-tune the effect model, as the nuisance outcome model is used the same way as in our \DMLForecaster.

\paragraph{Real-World Data}
Both nuisance models 
have an input dimension of 66, with multiple attention layers in
encoder and decoder, and 22 attention heads. 
The batch size for all nuisance models is $1200$ time series windows, and each had a learning rate scheduler of the form $ 
lr_{n} \mapsto \exp(\alpha)\cdot lr_{n} := lr_{n+1}\; ,
$
where $lr_n$ is the learning rate in the $n^{th}$ training step.

For the effect model we use twice the batch size as for the nuisance models (2400) which is due to the cross-fitting procedure (see \cref{sec:effect_model}). Moreover, we use a simple learning rate scheduler of the form
$$lr \mapsto \dfrac{lr}{\sqrt{n+1}}:=lr_n$$
where $lr_n$ is the learning rate after the $n^{th}$ training step and $lr$ is the initial learning rate.
\subsection{Experiments on Synthetic Data}
\begin{table*}[ht]
\centering
\begin{tabular}{lllll|ll}
\toprule
 & \multicolumn{2}{c}{MAE} & \multicolumn{2}{c|}{MSE} & MAE effect & MSE effect \\
Model type & Off policy & On policy & Off policy & On policy &  &  \\
\midrule
\Baseline & 16.3±0.5 & 11.5±0.4 & 745.7±38.6 & 490.6±19.4 & 45.8±1.0 & 3350.4±164.6 \\
DML & \bf{12.4±0.7} & \bf{10.0±0.7} & \bf{658.6±40.6} & \bf{472.9±33.9} & 25.0±1.7 & 1743.9±187.7 \\
\midrule
DML-no cf & \bf{12.4±0.7} & 10.1±0.7 & 663.2±49.0 & 473.6±33.4 & \bf{22.9±2.7} & \bf{1458.2±212.9} \\
sDML & 20.5±0.5 & 11.0±0.7 & 922.3±34.7 & 501.8±36.0 & 89.1±0.7 & 10356.5±251.9 \\
sDML-no cf & 20.5±0.6 & 11.0±0.7 & 919.4±37.2 & 499.8±35.7 & 89.5±1.1 & 10424.0±219.8 \\
\bottomrule
\end{tabular}
\caption{Error metrics predicting out-of-sample demand in study of 4500 simulated articles. See text for further details.}
\label{tab:sim-results}
\end{table*}
We start by providing a high-level overview on the construction of synthetic data to evaluate our approach in a controlled setting (see 
~\cref{app:simulation} for further details on the data generating process).

We simulate entire life cycles (100 weeks, typical in the online fashion industry) of around 4500 stock keeping units. Demand in a given week $t$ of article $i$ $q_{i,t}$ is a linear function of price $p_{i,t}$ and an article specific factor $e_i$ (treatment effect) as well as a base demand $\bsdem$, i.e.
\begin{equation}
    q_{i t} = \bsdem +  p_{i t} \, e_i
\end{equation}
Here treatment effects $e_i$ are article dependent, but constant over time. Note however that elasticity will not be constant over time:  $\epsilon_{i, t} = \frac{e_i p_{i t}}{\bsdem +  p_{i t} \, e_i}$.

The base demand $\bsdem$ is the product of two time dependent components: a noisy trend $\tau_{i t}$ that either leads to a linear increase/decrease of demand over the course of the article life cycle, and a seasonality term $s_{i t}$: 
\begin{equation}
\label{eq:base_demand}
    q_{i t}^{b} = (\tau_{i t}\cdot s_{i t} + 1)\cdot (c_{i}\,\lambda_{i t} + \eta_{i t}).
\end{equation}
The seasonality has a period of 30 weeks with an article-dependent phase shift in order to simulate different season types.
In addition, we scale our time-dependent component with an article-specific factor $c_{i}$ as well as independent additive- and multiplicative noise ($\eta_{i t}$ and $\lambda_{i t}$ respectively). Note, because of the product form in~\eqref{eq:base_demand}, our simulated noise is scale dependent on the base demand.
\begin{figure}
    \centering
    \includegraphics[width=\linewidth]{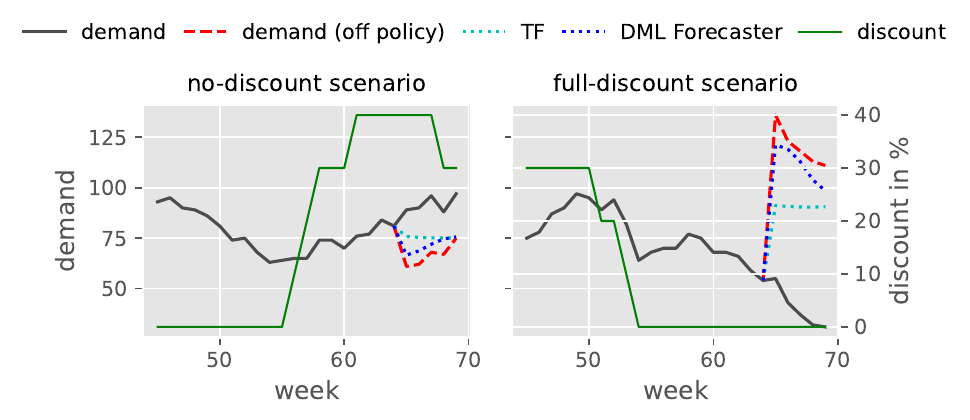}
    \caption{Synthetic demand time series (black), the associated realized discount (green) and off-policy forecasts for \DMLForecaster (blue) as well as \Baseline (cyan).}
    \label{fig:example-unit}
\end{figure}
Given our recipe to generate demand, we initialize the simulation for each article $i$ at week $t=0$ by setting an initial stock and price $p_{i 1}$. Our goal is to clear the given stock at $t=99$, the season end. We therefore simulate a pricing policy that, at any given week $t>3$, computes the average demand over the past four weeks ($t-1, t-2, \dots, t-4$). We then use this estimate to predict the week number at which the given article $i$ will run out of stock by mere linear extrapolation. If we estimate to clear our stock after $t=99$, we decrease our price by 10\% w.r.t. our base price $p_{i 0}$ in order to set $p_{i t}$. Conversely, if we expect to clear stock before season end, we increase $p_{i t}$ by 10\%.\footnote{We will open-source the data and data generation process (implemented in~\citep{gluonts_jmlr} as part of the publication.}  

Importantly, using such a pricing strategy, treatment is confounded by the long-term seasonal pattern of simulated demand (see example time series in \cref{fig:example-unit}). This leads to higher article discounts when the seasonal component of the  simulation is low (\cref{fig:example-unit}, left panel) and lower discounts when seasonal demand is high (\cref{fig:example-unit}, right panel).
\begin{figure}
    \centering
    \includegraphics[width=0.8\linewidth]{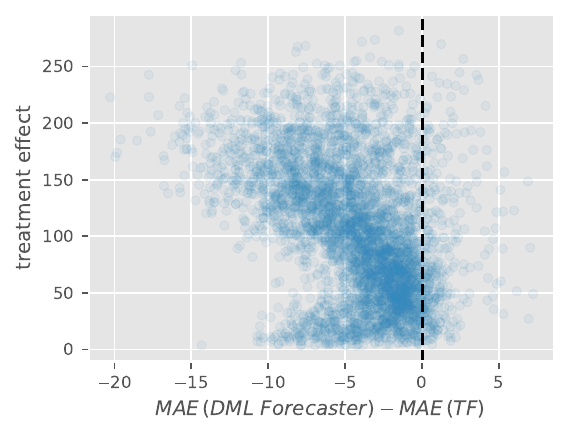}
    \caption{The improvement of the \DMLForecaster (the more negative on the x-axis the more improvement) over \Baseline increases with more elastic articles.}
    \label{fig:scatter}
\end{figure}
 We chose a total of four different periods for training: weeks 20-65, 30-75, 40-85, as well as weeks 50-95 and evaluate alternative methods on the five weeks that follow each training interval (weeks 66-70, 76-80, 86-90 and weeks 96-100 respectively). The  evaluation consists of two parts: \emph{on-policy} evaluation, where we predict demand under the pricing policy used in the simulation, as well as \emph{off-policy} evaluation, where we predict demand under five alternative discount levels that range from 0-50\% discount (w.r.t our initial price $p_{i 0}$). We repeat training and inference of all models five times to compute empirical standard deviations. In \cref{fig:example-unit} we show off-policy predictions of the \DMLForecaster and \Baseline when applying 0\% discount to weeks 65-70 (left panel) and the full discount (50\%) respectively (right panel). 

In addition to computing the standard metrics MAE and MSE on on- and off-policy ground truth, we also report how accurately our methods predict the treatment effect (MAE effect and MSE effect) -- as this parameter is directly modeled and inferred by each of the alternative models.

\paragraph{Model adaptations for this simulation study}
Because we deviate from the real-world constant elasticity assumption, we adapt the head effect model as introduced in \cref{eq:effecthead} accordingly, i.e. our final output is computed as
\begin{equation}\label{eq:simeffecthead}
\widehat{q^t}=\widetilde{q_t} + \psi(z) \cdot (d_t-\widetilde{d_t})
\end{equation}
where $\widetilde{q_t}$, $d_t$, and $\widetilde{d_t}$ are defined as in~\cref{eq:effecthead}. 

We change \Baseline accordingly, i.e. the head is computed as in \cref{eq:simeffecthead}, but we set $\widetilde{d_t}=0$.
\paragraph{Results}
We find that our \DMLForecaster (\DML) consistently outperforms  \Baseline when it comes to predicting demand under off-policy price changes (see \cref{tab:sim-results}). On policy, the difference between both models is not significant\footnote{W.r.t. computed empirical standard deviations}. Furthermore, we find that the advantage of using \DML over \Baseline is increasing with the size of the treatment effect (\cref{fig:scatter}).

\cref{tab:sim-results} further contains an ablation study which shows the results of two-stage methods that only learn a nuisance model for predicting demand (sDML and sDML-no cf) and find that they generally perform inferior in off-policy settings and in terms of estimating the effect of price changes.

With this simulation setup, we cannot confirm the benefit of using cross-fitting as the performance of \DML and \DML-no cf as well as (sDML and sDML-no cf) does not differ significantly across all error metrics we report here (\cref{tab:sim-results}). We have three explanations for this. First, that the ensembling in~\eqref{eq:ensemble} removes the benefit of cross fitting. Second, that some residual confounding may be large enough to obscure the benefits of cross fitting. Last, cross fitting is implemented to improve efficiency and statistical power. We may have enough data to fit the model.

\subsection{Cyberweek: Off-policy Discount Increase}
\begin{figure}
    \centering
    \includegraphics[width=0.6\linewidth]{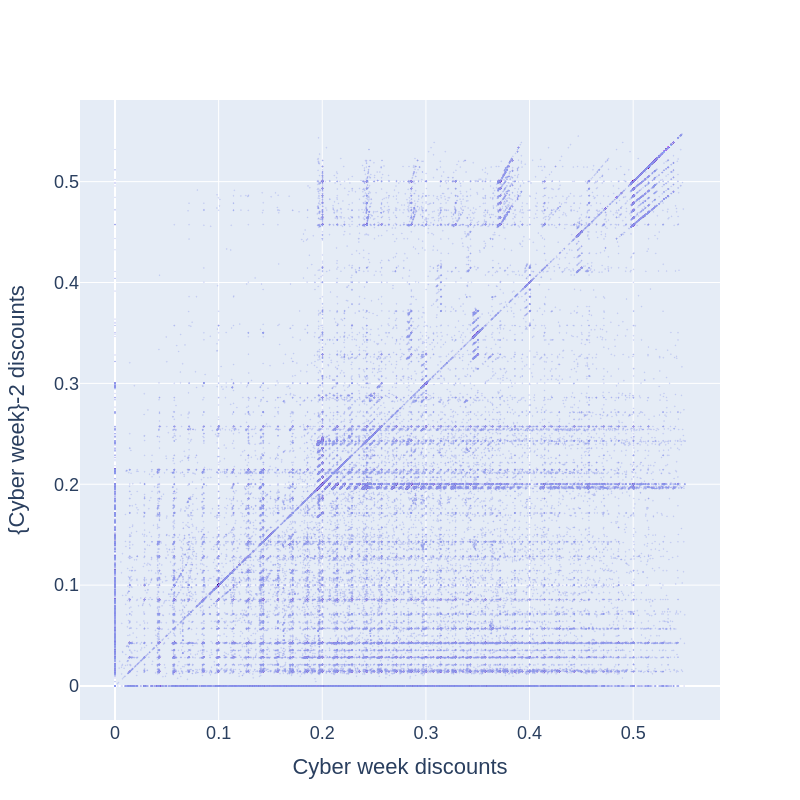}
    \caption{A scatter plot of discounts for articles on cyber week 2022 vs. two weeks prior. Each point represents a single article, and the units on the axes are the ratio of discount, with 0 being no discount and 1 being full discount.}
    \label{fig:cyberweek}
\end{figure}

\begin{table*}
  \begin{tabular}{cc|cccc|cccc}
\toprule
\multicolumn{1}{c}{}    
&
\multicolumn{1}{c}{}    
&&
\multicolumn{3}{c}{Off policy}    
&&                                           
\multicolumn{3}{c}{On policy} 
\\
\cmidrule(l){4-6}\cmidrule(l){7-10}     
Target Date & Metric &&\DMLForecaster & \Baseline & SARIMAX && \DMLForecaster & \Baseline & SARIMAX\\
\midrule
21-11-2022 & Demand Error && \textbf{61.48} & 80.03 & 81.33 && 60.00 & \textbf{54.73} & 81.08\\
23-11-2020 & &&\textbf{65.94} & 88.05  & 78.98 && 62.50 & \textbf{57.37}  &78.75\\
25-11-2019 & && 63.61 & \textbf{63.18} & 73.03 && \text{61.85}  &\textbf{57.29} & 69.59\\
\midrule
21-11-2022 & MAE &&\textbf{7.739} & 10.14 & 9.99 && 7.606 & \textbf{6.931} & 9.96\\
23-11-2020 & &&\textbf{12.92} & 17.39 & 14.82 && 12.39 & \textbf{11.34} & 14.78\\
25-11-2019 &  && 12.68 & \textbf{12.52} & 14.31 &&12.19 &\textbf{11.40} & 13.94\\
\midrule
21-11-2022 & MSE &&\textbf{2047} & 2540 & 2225 &&\textbf{1903} &1940 & 2196\\ 
23-11-2020 & && \textbf{5018}& 7361 & 4891 &&5092 &\textbf{5032} & 5062\\
25-11-2019 &  && \textbf{5075} & 5630 & 4348 &&5446 &5448 & \textbf{4472}\\
\bottomrule
\end{tabular}
\label{tab:cwresults}
\caption{Table of metrics for experiment dates considering \Baseline and \DMLForecaster for both off and on policy evaluation for cyberweek. All models were trained with an $L1$ loss function. Metrics read from the test epoch output.}
\medskip
\end{table*}
One way to test the price response of the models considers certain time periods where the discount policy follows a shifted distribution. In particular, \emph{cyber week} is such a yearly event when many articles have discounts that are much higher than normally seen during the year. For example, in \cref{fig:cyberweek} we look at the difference in discounts in cyber week 2022 versus two weeks prior at the article level and we see a general right-ward shift, which indicates the general increase in discounting on this special week.
Naturally, similar discount ranges occur during the same week in years prior, so it would not be an interventional test if each model saw these discount-time distributions in training. In order to test our hypothesis, we therefore discard cyber week, cyber week $-1$, and cyber week $+1$ from our training data and replace them with a set of 3 consecutive weeks that are resampled from the same article. We refer to data sets with discarded and replaced weeks as \textit{off-policy} and to data sets without this replacement as \textit{on-policy}.

We validate each model on 2021 cyber week, both on- and off-policy, and test each model on 2022, 2020, and 2019 cyber weeks on- and off-policy. Each experiment has a forecast horizon of cyber week and cyber week $+1$, while training on two years of article histories up until cyber week $-1$. The number of articles at inference time was $410,500$ for 2022, $208,212$ for 2020, and $144,980$ for 2019. We give an in depth qualitative description of our data in \cref{app:realworld}.

\paragraph{Results}
As shown in \cref{tab:cwresults}, we find that the \Baseline has a slight advantage when it comes to predicting demand on-policy, whereas the \DMLForecaster yields better results in the off-policy setting, particularly on the MSE. The SARIMAX model is consistently outperformed by both Transformer-based methods. 

In addition, we evaluate our methods on control dates not affected by cyber-week sales events (\cref{tab:cwcontrol}), and we show how \DML improves over \Baseline w.r.t. the degree of discount change  (\cref{fig:errorplotcw}).

\section{Related Work}\label{sec:related_work}
The wider areas of forecasting and causal inference, especially in a pricing context, are well established fields, but typically studied in isolation. 

Forecasting with transformers has received considerable attention in the literature in both academic and industrial research~\citep{haoyietal-informer-2021,kunz2023deep,eisenach2020mqtransformer,lim2021} and they are generally acknowledged to work well for real-world data sets such as the ones we consider in Section~\ref{sec:experiments}. Our approach is generic in the sense that it would work with other transformer-based methods (or indeed, other forecasting methods as long as they allow for the incorporation of covariates). We chose a specific architecture for the ease of implementation and customization to the pricing use case~\citep{kunz2023deep}.

In econometrics, demand estimation via price elasticities is of central interest~\cite{deaton1980economics,fogarty2010demand,hughes2008evidence,defusco2017interest}. Often however, forecasting methods are ignored as the focus is understanding how demand changes when prices or policies change. Recent work has shown how using forecasting algorithms to complement existing econometric techniques can improve causal inference~\cite{goldin2022forecasting}. We do the opposite, using causal inference methods to improve forecast estimates.  

\emph{Causal forecasting}, that is, the intersection of causal inference and forecasting, is typically only mentioned briefly in standard forecasting textbooks (see e.g.,~\citep{hyndman2017forecasting}). Similarly, research in causal forecasting is limited to the best of our knowledge. There are some notable exceptions, including the above example using forecasting algorithms to improve causal inference estimates~\cite{goldin2022forecasting}. For example, \citet{vankadara2022causal} provide a theoretical framework for differentiating the causal from the statistical risk in forecasting. Our work is more pragmatically oriented in the sense that we do not make assumptions on causal sufficiency (but also do not obtain theoretical guarantees) and rather focus on the empirical validity and evaluation of our approach. 

The wider area of estimating counterfactuals is well-studied especially in a medical setting with discrete treatments. \citet{pmlr-v162-melnychuk22a} recently provide a transformer-based approach for estimating counterfactual outcomes in a discrete treatment setting with medical data. They provide an end-to-end training procedure of three sub-models instead of the multi-stage approach presented here. While a multi-stage approach introduces inefficiencies, the joint loss in~\citep{pmlr-v162-melnychuk22a} relies on adversarial learning of multiple objectives not in the same domain or scale which is notoriously hard to tune. Our work provides the estimation of a causal parameters of interest, the price elasticity, which~\citep{pmlr-v162-melnychuk22a} doesn't yield. Similarly, ~\citet{johansson2016} predict patient
outcomes over simulated data using a deep neural network approach that corrects
for the treatment bias by a given patient's medical history. ~\citet{bica2020} extend
this work to a longitudinal setting, estimating counterfactual patient outcome
timeseries while accounting for time-varying confounders. A notable exception
is~\citep{pawlowski2020deep}, where factual
information of the period of interest is used in the model in order to
compute counterfactuals. 
Approximating counterfactuals by interventional distributions~\cite{johansson2016,bica2020,pmlr-v162-melnychuk22a} has the
advantage that the resulting methods are by design applicable to interventional
settings like ours. Conversely, our proposed approach may also be used to
estimate counterfactual outcomes.

\section{Conclusion and Future Work}\label{sec:conclusion}
We presented a causal forecasting method in a pricing context via \DML. Our model relies on state-of-the-art transformer-based forecasting models and, by incorporating \DML, allows for off-policy estimations and a better causal effect estimation than purpose-built, but causally unaware forecasting methods. The evaluation of such forecasts is notoriously difficult and we provide synthetic data as well as natural experiment data for such evaluations. 

Future work should include a probabilistic treatment and the incorporation of inverse propensity scores~\citep{lim18}, a more flexible outcome model as well as the inclusion of multi-variate forecasting models.

\bibliography{dml.bib}
\clearpage

\appendix

\thispagestyle{empty}
\onecolumn

\title{Causal Forecasting for Pricing: supplemental material}
\author{}
\date{}
\maketitle

\section{Differences Between Cross-fitting and Sample-splitting}\label{app:diff_sample_cross}

Cross-fitting, as deployed in the \DMLForecaster, has a primitive method called \emph{sample-splitting}. This is a simple version of the cross-fitting DML estimation described in literature~\citep{chern2017}. With the \DMLForecaster we split the training data into two randomized subsets (which we refer to as \emph{even} and \emph{odd}) as in cross-fitting. However, there is only one of each outcome and treatment model. The nuisance models are trained on the even part of the dataset, while the effect model is trained on the odd part with partial ground truth in~\eqref{eq:effecthead} provided by the nuisance models' inference on the odd part. At inference time we use the nuisance and effect models to forecast on the full assortment.

An issue with sample-splitting is that we only use half the dataset for inference. This means that this method is less efficient since it does not use all of the data for training each of the nuisance or effect model.

In our use case there is no shortage of data, and so it becomes a question as to whether this efficiency is really needed. In addition, cross-fitting would add increased training and inference time over sample-splitting. We would like to understand the trade off between the two methods, and therefore designed a small experiment to test the sample-splitting DML model against the cross-fitting DML model, where the former is designed as in the first paragraph in this sub-appendix.

We tested the sample-splitting DML model on the three cyber week start dates, both in and out of sample, and on the control dates in sample in order to compare with the cross-fitting DML model. The hyperparameters for each sub-model in the sample-splitting DML model are taken from the tuning study where cross-fitting was in place. We also measured training time from the tensorboard logs. 

The cross-fitting \DMLForecaster performed slightly better than the sample-splitting \DMLForecaster off-policy. For on-policy, the sample-splitting model performed slightly better than the cross-fitting model on two of the cyber week dates and one of the control dates. On the remaining dates it performed slightly worse. The training time for cross-fitting was significantly longer than that for sample-splitting. See \cref{tab:ssexp} for the results.

\section{Derivation of \cref{eq:elasticity}}\label{app:derivation_demand_eq}
We make the assumption that demand depends negative monotonically on price for a specific article at a specific time, with all other factors being held constant. Further, we assume that elasticity is constant with regard to price, demand, and time. We can then treat \cref{eq:elasticity} as an equality of differential forms on the positive real line
\begin{eqnarray*}
\frac{dq}{q}  =  \epsilon \frac{dp}{p}.
\end{eqnarray*}
If we integrate both sides over some interval $[p_0, p_1]$, 
\begin{eqnarray*}
    \int_{p_0}^{p_1} \frac{d(q(p))}{q(p)}  =  \epsilon \int_{p_0}^{p_1} \frac{dp}{p} .
\end{eqnarray*}
We can make the substitution of $q(p) = q$ in the left hand side, and get
\begin{eqnarray*}
    \int_{q_0}^{q_1} \frac{dq}{q}  =  \epsilon \int_{p_0}^{p_1} \frac{dp}{p}
\end{eqnarray*}
where $q_i = q(p_i)$ for $i=0,1$. Thus, we get
\begin{eqnarray*}
    \log(q_1)-\log(q_0) = \epsilon \big(\log(p_1)-\log(p_0)\big).
\end{eqnarray*}
Exponentiating, we arrive at the result:
\begin{eqnarray*}
q_1=q_0\Big(\frac{p_1}{p_0}\Big)^\epsilon.
\end{eqnarray*}
\section{Elastiticity-based Forecasts}\label{app:TWFE}

A standard approach to compute demand functions uses estimated elasticities, $\varepsilon$ (see for instance \citet{varian2014intermediate}, \citet{deaton1980economics} or \citet{phillips2021pricing}). 

Here, we use estimated elasticities for different groups of articles and past observed demand in order to predict demand for a given week $t$, i.e.
\begin{equation*}
    \hat{q}_{i, t}=q_{i, t-1}\left(\frac{1-d_{i, t}}{1-d_{i, t-1}}\right)^{\varepsilon}.
\end{equation*}
Due to hidden confounding (e.g., seasonality and advertisement campaigns), na\"ive regression of $\log(q_{i, \cdot})$ onto $\log(1-d_{i,\cdot})$ generally results in biased estimates of the elasticities. To account for this, we use a two-way fixed-effects Poisson regression model that is defined as follows:
\begin{equation}\label{eq:fixed_effects}
    \log\big(\expec{q_{i, t}}\big)=\varepsilon\log(1-d_{i, t})+u_i+c_t.
\end{equation}
Here, the parameter $u_i$ is the \emph{article-specific} effect and $c_t$ is the \emph{week-specific} effect. 
This model is fitted with the standard \emph{within estimator} using the R package \texttt{fixest} \cite{fixest}. 

\section{Hyperparameters and Hyperparameter Tuning}
\label{app:hyperparameters}

Hyperparameter tuning was carried out using the Bayesian search algorithm provided with the python package \emph{optuna}~\citep{akiba2019optuna}. We provide a full overview of the selected hyperparameters for the \Baseline model in~\cref{tab:hyperp_baseline} and for the \DMLForecaster in~\cref{tab:hyperp_dml}. A notable difference between \Baseline and \DMLForecaster is that we used RAdam in each DML sub-model and AdamW for the \Baseline model. For tuning the effect model, we trained the nuisance models with optimal hyperparameters and saved a checkpoint of their weights, and then tuned the effect model for its set of parameters without further training of the nuisance models.

\section{Simulation Study}
\label{app:simulation}
In the following, we give more details about the generation of our synthetic data set. Note that, in the main manuscript, we keep our presentation concise and on a high level by omitting the weighting we use for individual components and using a different (but equivalent) parameterization of the noise model.

We simulate a total of 4467 stock keeping units over a period of 100 weeks (i.e $t \in \{0,1, \dots ,99\}$). Demand in a given week $t$ of article $i$ ($q_{i,t}$) is a linear function of price $p_{i,t}$ and an article-specific factor $e_i$ (treatment effect) as well as a base demand $\bsdem$, i.e.
\begin{equation}
    q_{i t} = \bsdem +  p_{i t} \, e_i
\end{equation}
The base demand $\bsdem$ is the product of two time dependent components: a noisy trend $\tau_{i t}$ that either leads to a linear increase/decrease of demand over the course of the article life cycle, and a seasonality term $s_{i t}$: 
\begin{equation}
\label{eq:base_syn}
    q_{i t}^{b} = (0.15 \cdot \tau_{i t} + 0.25 \cdot s_{i t} + 1)\cdot c_{it}.
\end{equation}
$c_{i t}$ is the article-specific contribution that consists of two sub components $a_{i t}$ and $b_{i t}$:
\begin{eqnarray}
c_{i t} = 0.05\cdot a_{i t} ^2 + 0.25 \cdot a_{i t} + 0.5 \cdot b_{i t}
\end{eqnarray}
where 
\begin{equation}
    a_{i t} = \alpha_{d(i)} + \epsilon_{i t}
\end{equation}
and
\begin{equation}
  \alpha_{d} \sim \mathcal{N}(10,\, 3^2)
\end{equation}
is sampled once for each category $d \in \{1, 2, \dots , 45\}$. Furthermore $\epsilon_{i t}$ is independent noise drawn from $\mathcal{N}(0,\, 1)$ and $d(i)$ is a (random) mapping that assigns article $i$ to one of a total of 45 categories.

The contribution of $b_{i t}$ is computed analogously to $a_{i t}$ using a different setting of hyperparemeters and a total of 15 categories:
\begin{eqnarray}
b_{i t} &=& \beta_{k(i)} + \psi_{i t}, \, \psi_{i t} \sim \mathcal{N}(0,\, 5^2)\\
  \beta_{k} &\sim&\mathcal{N}(300,\, 50^2), \, k \in \{1, 2, \dots, 15\}.
\end{eqnarray}

The treatment effect $e_i$ in \cref{eq:base_syn} depends on a random component as well as an article-specific component:
\begin{eqnarray}
    e_{i} = e^{(b)}_i  \cdot 0.15 \cdot \bar a_i, \,\, e^{(b)}_i \sim \text{max}(1.3, \mathcal{LN}(0.75,\, 0.125^2))
\end{eqnarray}
where $\bar a_i = \frac{1}{100} \sum_{t=0}^{99} a_{i t}$

Furthermore we chose our initial price $p_{i 0}$ such that we avoid $q_{i t} < 0$ at any week $t$:
\begin{equation}
    p_{i 0} \sim \mathcal{N}\left(\frac{\bar q_i}{3}, \, \left(\frac{\bar q_i}{1.5}  \right)^2 \right),
\end{equation}
where $\bar q_i = \frac{1}{100} \sum_{t=0}^{99} \bsdem$. We apply additional filtering steps to exclude articles that have negative demand from our synthetic data set.
The seasonal component $s_{i t}$ in \cref{eq:base_syn} is a sine function with a period of 30 (weeks) and article-dependent shifts (season types) that are tied to our categorical variable $k$. In particular, we subdivide the values of $k$ evenly into six subgroups and sample season shifts for each group uniformly over all integers in the interval $[-15, 15]$.

Lastly, our trend component $\tau_{it}$ follows a (noisy) linear function, i.e.
\begin{equation}
    \tau_{it} \sim \mathcal{N}(t \cdot \gamma_i,\, \sigma_{\tau_i}^2)
\end{equation}
where
\begin{equation}
    \gamma_i \sim \mathcal{U}([-0.02, 0.02])
\end{equation}
and
\begin{equation}
    \sigma_{\tau_i} \sim \mathcal{U}([0, 0.15]).
\end{equation}

With this demand model in place we set our initial stock $z_0$ such that we clear our simulated inventory in week $t=99$ at an average discount rate of 14\%, i.e
\begin{equation}
    z_0 = \frac{1}{100}\sum_{t=0}^{99} \bsdem \cdot (1-0.14)p_{i 0}e_i. 
\end{equation}
Moving along, we compute demand for the first four weeks, i.e $q_{i t}$ for $t \in \{0, 1, 2, 3\}$ keeping the price constant ($p_{i 0} = p_{i 1} = p_{i 2} = p_{i 3}$). For all weeks $t$ we update our stock accordingly:
\begin{equation}
    z_{t+1} = z_{t} - q_{i t}
\end{equation}
At any given week $t \in \{4,5, \dots 100\}$, we compute the expected number of weeks until we run out of stock ($m_{t}$), via a basic linear extrapolation:
\begin{equation}
    m_{t} = \frac{4 z_{t}}{\sum_{t_j=t-3}^{t} q_{i t_j}},
\end{equation}
which we can use to compute the so-called stock coverage $w_t$:
\begin{equation}
    w_{t} = \frac{m_{t}}{100-t}.
\end{equation}
A value of $w_{t} > 1$ implies that demand is too low in order to clear stock at season end, and conversely, a value of $w_{t} < 1$ would lead to left over stock after our period of 100 weeks. 

Our pricing policy is set up in order to steer $w_{t}$ toward 1 for all $t \in \{4,5, \dots, 99 \}$. In particular, we define a total of six discount steps $d(j_t) = j_t\cdot0.1$ for $j_t \in \{0, 1, \dots, 5\}$ for a given week $t$ and adjust our discount step according to the following probabilistic rule:
\begin{equation}
    j_t = \begin{cases}
j_{t-1} + 1: w_{t} > 1, &\lambda_{t i} > \frac{1}{w_t}\\
j_{t-1} - 1: w_{t} < 1, &\lambda_{t i} > w_t \\
j_{t-1}:&\text{ otherwise},
\end{cases}
\end{equation}
where $\lambda_{t i} \sim \mathcal{U}([0,1])$. We then update our price $p_{it}$ in order to compute demand $q_{i t}$ via \cref{eq:base_demand} as follows:
\begin{equation}
    p_{it} = p_{i 0} \cdot (1-d(j_t)).
\end{equation}

We give an overview of the synthetic data set and how we derive features from it in \cref{tab:sinfeatures}.

\begin{table}
  \begin{tabular}{l|l|l}
Feature & Data type & Notes\\
\hline
\specialrule{0em}{.25em}{.1em}
\multicolumn{3}{c}{Dynamic Features} \\
\hline
Demand & Integer $\geq 0$ & simulated demand\\
Discount & Float & range between 0 and 0.5\\
Stock & Integer $\geq 0$ & available stock \\
Week number & Integer $\geq 0$ & week number (embedded) \\
Positional Encoding & Float(x17) & positional encoding dimensions\\
\hline
\specialrule{0em}{.25em}{.1em}
\multicolumn{3}{c}{Static Features} \\
\hline
$d$ & Categorical & embedded via a learned embedding \\
$k$ & Categorical & embedded via a learned embedding\\
Promotion & Binary & noise: not having an effect on demand\\
$p_0$ & Integer & undiscounted price of the article\\
\end{tabular}
\medskip

\caption{Overview of syntetic dataset and its usage with the \DMLForecaster and \Baseline}\label{tab:sinfeatures}
\medskip
\end{table}

\section{Details on Experiments for Real World Data}
\label{app:realworld}
\subsection{Qualitative description of our data}
Our data consists of sales and other recorded properties of fashion articles that
were sold at some point in the past via the retailers online shop. We
refer to a single article as Stock Keeping Unit (SKU), and in the following, we
consider the so-called config Stock Keeping Units (cSKUs) that group the same
articles of different sizes. Thus, all data presented here
is agnostic of article size and we ignore effects that are the result of an
cSKU being available in a limited number of sizes at some point in its life
cycle.

Each cSKU comes with its associated history of weekly-aggregated observations
and features. Depending on the context, we use the shorthand \emph{cSKU} to also refer
to a given article's history. We give more detail on the recorded history and
derived features available for each cSKU in \cref{tab:datafeatures}.

\begin{table}
  \begin{tabular}{l|l|l}
Feature & Data type & Notes\\
\hline
\specialrule{0em}{.25em}{.1em}
\multicolumn{3}{c}{Dynamic Features} \\
\hline
Sold Items & Integer $\geq 0$ & sold items before return\\
Discount & Float & range between 0 and 0.7\\
Stock & Integer $\geq 0$ & available stock for a given cSKU \\
Week number & Integer $\geq 0$ & iso calendar week number(embedded) \\
Day in year & Integer $\geq 0$ & number of days from January 1st (embedded) \\
Days from Easter & Integer $\geq 0$ & number of days from Easter (embedded) \\
Positional Encoding & Float(x17) & positional encoding dimensions\\
\hline
\specialrule{0em}{.25em}{.1em}
\multicolumn{3}{c}{Static Features} \\
\hline
Brand & Categorical & embedded via a learned embedding \\
Commodity group & Categorical (x5) & hierarchical category groups (embedded)\\
Season type & Categorical & season type of article (embedded)\\
Black price & Integer & undiscounted price of the article\\
\end{tabular}
\medskip

\caption{Overview of real-world dataset and its usage with the \DMLForecaster and \Baseline}\label{tab:datafeatures}
\medskip
\end{table}

\begin{table*}[t]
	\begin{center}
		\begin{scriptsize}
			\begin{sc}
				\begin{tabular}{lccccc}
    \toprule
Characteristic &  Synthetic & Cyber week 2019 & Cyber week 2020 & Cyber week 2022\\
\midrule
No. Time Series &$4,467$& $144,980$ & $208,212$ & $410,500$\\
Time Granularity &weekly & weekly & weekly & weekly\\	
Avg. Length of Time Series &$100$ & $104$ & $100$ & $75$\\
forecast horizon length &$5$ & $2$ & $2$ & $2$\\ 
                \bottomrule
                    \end{tabular}
                \end{sc}
                \caption{High-level characteristics of data sets.}
                \label{tab:summary_ds}
            \end{scriptsize}
        \end{center}
\end{table*}

We use one-hot encoding to compute a high-dimensional numeric vector for each
categorical feature that we pass through an embedding layer to obtain a
low-dimensional representation.  Similarly, we use embeddings for our ordinal
features (Isoweek number, Days from January first, and Days from Easter). Note, that all
embedding layers are an integral part of the neural networks we present here.
Thus, during training, we update the parameters of the embedding layers as part of
the same gradient update that we use to optimize the remaining weights of
each model.

We treat discount as a continuous variable -- even though inventory managers
typically reduce prices by increments of five percent relative to some baseline
(black price). In practice, we need a higher resolution as discounts are
recorded as weekly averages. Depending on the time a discount is
updated this can result in rather arbitrary decimals. 
For instance, if the discount for a given fashion item is increased from 20\% to 25\% in
the middle of a given week, we would record a an aggregated discount level of
22.5\%.

\subsection{Further Results}
The main cyber-week results are given in \cref{tab:cwresults} with the comparison to the control dates given in \cref{tab:cwcontrol}. On the 2022 and 2020 cyber weeks, the \DMLForecaster performed substantially better across all metrics when compared to the \Baseline model. As we expected, the \Baseline performed mildly better than the \DMLForecaster on policy for the cyber weeks. On the 2019 cyber week, the \DMLForecaster and \Baseline models performed similarly in the off policy test, and the \Baseline was once again better on policy.

To understand the magnitude of the results with regard to the experiment design, we look at \cref{tab:cwcontrol}. Indeed, the degree of change in the error metrics for the \Baseline is not drastic when moving from on policy to off policy for each specific date, when compared to the 2022 and 2020 cyber week dates. The degree of change in the error metrics for the \DMLForecaster when considering the on-to-off policy shift is similar to the cyber week dates.
\begin{figure}
    \centering
    \includegraphics[width=0.5\linewidth]{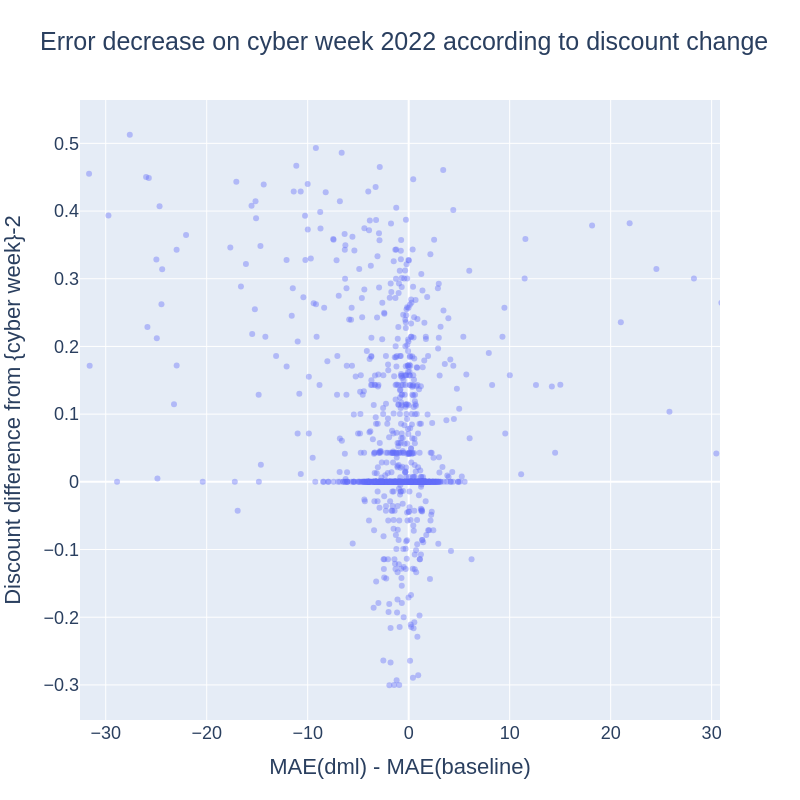}
    \caption{A sample of the difference in forecasting error for the \Baseline vs. \DMLForecaster on cyber week 2022, measured on the off-policy experiment.}
    \label{fig:errorplotcw}
\end{figure}
\begin{table}
\centering
\begin{tabular}{cc|ccc|ccc|}
\toprule
\multicolumn{1}{c}{}    
&
\multicolumn{1}{c}{}    
&&
\multicolumn{2}{c}{Off policy}    
&&                                           
\multicolumn{2}{c}{On policy} 
\\
\cmidrule(l){4-5}\cmidrule(l){7-8}     
Target Date & Metric &&DML& Baseline    && DML & Baseline      \\
\midrule
& Demand Error&& & && &\\
25-04-2022 & && \textbf{51.03} & 57.84  && \textbf{46.55} & 48.59 \\
06-06-2022 & && \textbf{53.76} & 58.61  && 47.84 &\textbf{45.61} \\
10-10-2022 & &&\textbf{52.48} & 56.18  && 51.97 & \textbf{51.31} \\
\midrule
& MAE && & && &\\
25-04-2022 & && \textbf{8.935} & 10.15 && \textbf{8.065} & 8.478 \\
06-06-2022 & && \textbf{7.274} & 8.081 && 6.361 & \textbf{6.306} \\
10-10-2022 & &&\textbf{5.183} & 5.616 && 5.078 & \textbf{5.042} \\
\midrule
& MSE && & && &\\
25-04-2022 & && \textbf{1772} & 2144 && \textbf{1555} & 1861 \\
06-06-2022 & && \textbf{1009} & 1319 && \textbf{874.0} & 1007 \\
10-10-2022 & && \textbf{510.9} & 628.0 && \textbf{454.9} & 507.6 \\
\bottomrule
\end{tabular}
\caption{Table of metrics for control dates. We consider baseline and \DMLForecaster for both on policy and off policy evaluation. All models were trained with an $L1$ loss function. Metrics read from the output of the test epoch. \label{tab:cwcontrol}}
\medskip
\end{table}

\subsection{Further Ablation Study on Real World Data}\label{subsec:sampsplitmethods}

There are two key elements of the \DMLForecaster which are important in avoiding bias (see the Introduction of \citet{chernstructural} for a deeper discussion). The cross-fitting method allows $\sqrt{n}$-consistency in the linear effect case and prevents the effect model from overfitting. Thus, we can ask if this cross-fitting is necessary in our case: We run the \DMLForecaster without cross-fitting and compare to the original, on the start date of 21-11-2022, for both in and out of sample performance. More precisely, we use the even batches with the even nuisance models to provide inference for effect model training, and similarly for the odd version.

As a second experiment, there is a ``simplified" version of \DMLForecaster (\emph{sDML}), that does not orthogonalize the treatment function, see for example \cite[Equation 1.3]{chernstructural}. Equation \ref{eq:effecthead} simply becomes
\begin{equation}\label{eq:pmeffecthead}
\widehat{q^t}=\widetilde{q^t}\big(1-d_t\big)^{\psi(X)} \qquad 1\leq t \leq s,
\end{equation}
where $d_t$ is the desired discount. We test this model on the same start date of 21-11-2022.

As a final experiment, we test the sDML model without cross-fitting. 

The goal is that this ablation study will help explain the mechanism by which \DMLForecaster improves upon the \Baseline for out-of-sample forecasting.

There was not a significant difference between cross-fitting and no cross-fitting for either type of \DMLForecaster. The sDML model performed worse on- and off-policy for Demand Error and MAE, regardless of cross-fitting. This is explained by difference in error between the final output and the outcome model. The effect model in the regular \DMLForecaster corrected the Demand Error by $6.65$ resp. $2.57$ for off- resp. on-policy over the Demand Error of the outcome model. However, For the sDML model, this correction was significantly less for off policy, and the effect model even increased the error for on-policy. See \cref{tab:ablation} for the results.

\begin{table}
\centering
  \begin{tabular}{cc|cc|cc|cc}
  \toprule
\multicolumn{1}{c}{}    
&
\multicolumn{1}{c}{}    
&
\multicolumn{2}{c}{Off policy}    
&                                           
\multicolumn{2}{c}{On policy} 
&
\multicolumn{2}{c}{Train time}
\\
\cmidrule(l){3-4}\cmidrule(l){5-6}\cmidrule{7-8}     
Target Date & Metric &Cf& Ss    & Cf & Ss & Cf & Ss     \\
\midrule
& Demand Error& & & &\\
21-11-2022 &  & \textbf{61.48} & 62.46  & 60.00 & \textbf{58.41} & 2.55 & 1.63\\
23-11-2020 & &\textbf{65.94} & 68.67  &\textbf{62.50} & 64.24 & 1.17 & 0.71\\
25-11-2019 & &\textbf{63.61} & 67.52 &61.85  &\textbf{61.60} & 0.83 & 0.52\\
\midrule
25-04-2022 & &&  &  46.55 & \textbf{46.28} & 1.96 & 1.27\\
06-06-2022 & &&  & \textbf{47.84} & 50.26 & 2.11 & 1.36\\
10-10-2022 & &&  &\textbf{51.97} & 52.66 & 2.44 & 1.61\\
\bottomrule
\end{tabular}

\caption{Demand error for the cross-fitting \DMLForecaster compared to the sample-splitting \DMLForecaster on the cyber week and control week target dates. Cf = Cross-fitting, Ss = Sample-splitting. Training was done on an AWS Sagemaker instance of type G4dn.4xlarge. Training time is in hours.
\label{tab:ssexp}}
\medskip
\end{table}

\begin{table}
\centering
  \begin{tabular}{c|c|c|c|cc}
  \toprule
   \multicolumn{1}{c}{}    
&
\multicolumn{1}{c}{}    
&
\multicolumn{1}{c}{}    
&                                          
\multicolumn{1}{c}{} 
&                                          
\multicolumn{2}{c}{Effect error corr.} 
\\
\cmidrule(l){5-6} 
Model type & Metric & Off policy   & On policy & Off pol. & On policy   \\
\midrule
& Demand Error& &  \\
DML &  & \textbf{61.48}  & \textbf{60.00} & -6.65  &-2.57 \\
DML-no cf & &62.7   &60.92 & -5.43 &-1.95\\
sDML & &64.65 &64.34  & -3.48& +1.77\\
sDML-no cf & & 65.96 & 65.08&-2.17 & +2.51 \\
\midrule
& MAE& &  \\
DML &  & \textbf{7.739}  &\textbf{7.606}&& \\
DML-no cf & &7.883   &7.699 &&\\
sDML & &8.144&8.172 &&\\
sDML-no cf & &8.299 &8.279&& \\
\midrule
& MSE& &  \\
DML &  &\textbf{2047}  & 1903&& \\
DML-no cf & &2065   &1906 &&\\
sDML & &2067 &\textbf{1899} &&\\
sDML-no cf & &2091 &1904&& \\
\bottomrule

\end{tabular}
\caption{Metrics for ablation study for the target date of 21-11-2022. DML-no CF is DML with no cross-fitting, sDML is the simplified DML without the discount residual. Best Metrics for each block are bold. The effect error correction column is the difference in error between the outcome model and the output of the effect model (negative numbers indicate a reduction in error from the outcome model). Metrics read from the output of the test epoch. \label{tab:ablation}}
\medskip
\end{table}

\section{Details to hardware and library versions used}
We performed all experiments using either amazon web services' (aws) ml.g5.12xlarge instances (simulation study) or g4dn.4xlarge instances (experiments on cyber-week data). The experiments were conducted with PyTorch 2.0.0 and Python 3.10 installed.

\begin{table}
\centering
  \begin{tabular}{c|c|c|p{55mm}}
\specialrule{.15em}{.1em}{.1em}
Baseline Parameter & Cyberweek Data & Simulated Data & Notes\\
\specialrule{.15em}{.1em}{.1em}
Number of layers & $6$ & $13$ & number of residual blocks in the encoder and decoder\\
\midrule
Hidden dimension & $274$ & $51$ & number of hidden dims in ffn after each attention step\\
\midrule
Head hidden dimension & $164$ & $62$ & in the network that computes the normalized demand slopes\\
\midrule
Dropout & $0.41$ & $0.43$ & in all ffns and attention weights\\
\midrule
Learning rate & $0.0034$ & $0.0224$ & for the AdamW optimizer\\
\midrule
Weight decay & $0.037$ & $\num{1.4e-4}$ & regularization parameter\\
\midrule
Beta 1 & $0.8044$ & $0.8566$ & decay rate for computing moving average of gradient in the Adam optimizer\\
\midrule
Beta 2 & $0.6023$ & $0.9140$ & decay rate for computing moving average of gradient in the Adam optimizer\\
\midrule
Number of linear pieces & $2$ & $1$ & number of pieces for the demand curve\\
\midrule
Train epochs & $2$ & $23$ & \\
\midrule
Total parameters & 1.3M & 124K &\\
\specialrule{.1em}{.05em}{.05em}
\end{tabular}
\caption{Hyperparameters for the \Baseline with production data (cyberweek) and simulated data}\label{tab:hyperp_baseline}
\medskip
\end{table}

\begin{table}
\centering
  \begin{tabular}{c|c|c|p{65mm}}
\specialrule{.15em}{.1em}{.1em}
DML Parameter & Cyberweek Data & Simulated Data & Notes\\
\specialrule{.15em}{.1em}{.1em}
Outcome model\\
\midrule
Number of layers & $3$ & $5$ & number of residual blocks in encoder and decoder\\
\midrule
Hidden dimension & $130$ & $73$ & number of hidden dims in ffn after each attention step\\
\midrule
Dropout & $ 0.2$ & $0.1$ & in all ffns and attention weights\\
\midrule
Learning rate & $0.0096$ & $0.0088$ & for the RAdam optimizer\\
\midrule
Weight decay & $\num{1.4e-4}$ & $\num{5.4619e-9}$ & regularization parameter \\
\midrule
Beta 1 & $0.7096$ & $0.8566$ & decay rate for computing moving average of gradient in the Adam optimizer\\
\midrule
Beta 2 & $0.8585$ & $0.9140$ & decay rate for computing moving average of gradient in the Adam optimizer \\
\midrule
Gamma & $0.9993$ & $0.9388$ & Exponential decay of learn rate\\
\midrule
Train epochs & $1$ & $44$ & \\
\specialrule{.15em}{.1em}{.1em}
Treatment model\\
\midrule
Number of layers & $2$ & $2$ & number of residual blocks in encoder and decoder\\
\midrule
Hidden dimension & $160$ & $21$ & number of hidden dims in ffn after each attention step\\
\midrule
Dropout & $ 0.2$ & $0.1497$ & in all ffns and attention weights\\
\midrule
Learning rate & $\num{4.2e-4}$ & $0.0162$ & for the RAdam optimizer\\
\midrule
Weight decay & $\num{5.8e-4}$ & $\num{2.6e-09}$ & regularization parameter\\
\midrule
Beta 1 & $0.7884$ & $0.5977$ & decay rate for computing moving average of gradient in the Adam optimizer \\
\midrule
Beta 2 & $0.9536$ & $0.8740$ & decay rate for computing moving average of gradient in the Adam optimizer \\
\midrule
Gamma & $0.9995$ & $0.9549$ & Exponential decay of learn rate\\
\midrule
Train epochs & $1$ & $60$ & \\
\specialrule{.15em}{.1em}{.1em}
Effect model\\
\midrule
Number of layers & $5$ & $6$ & number of residual blocks in encoder and decoder\\
\midrule
Hidden dimension & $273$ & $43$ & number of hidden dims in ffn after each attention step\\
\midrule
Dropout & $0.412$ & $0.1750$ & in all ffns and attention weights\\
\midrule
Learning rate & $\num{1.07e-8}$ & $0.0491$ & for the RAdam optimizer\\
\midrule
Weight decay & $0.0375$ & $\num{5.75e-09}$ & regularization parameter\\
\midrule
Beta 1 & $0.6$ & $0.6405$ & decay rate for computing moving average of gradient in the Adam optimizer\\
\midrule
Beta 2 & $0.7044$ & $0.6749$ & decay rate for computing moving average of gradient in the Adam optimizer\\
\midrule
Train epochs & $1$ & $20$ & \\
\midrule
Total parameters & 1.2M & 118K &\\
\specialrule{.1em}{.05em}{.05em}
\end{tabular}
\caption{Hyperparameters for the \DMLForecaster with production data (cyberweek) and simulated data}\label{tab:hyperp_dml}
\end{table}

\end{document}